\documentclass[journal]{IEEEtai}

\usepackage[colorlinks,urlcolor=blue,linkcolor=blue,citecolor=blue]{hyperref}

\usepackage{color,array}
\usepackage{multirow}
\usepackage{booktabs}
\usepackage{graphicx}
\usepackage[justification=centering]{caption}
\usepackage{listings}
\usepackage[utf8]{inputenc}
\usepackage[most]{tcolorbox}
\usepackage{xcolor}
\usepackage{comment}
\usepackage{tcolorbox}
\tcbuselibrary{listings, breakable}
\definecolor{lightgraybox}{RGB}{245,245,245}
\definecolor{darkborder}{RGB}{100,100,100}
\usepackage{graphicx}
\usepackage{tcolorbox}
\usepackage{caption}  % For captionof
\usepackage{lipsum}   % (Optional) for filler text
\usepackage{cuted}    % Required to insert content across both columns

\usepackage[numbers,sort&compress]{natbib} % IEEE numerical style
\usepackage{amsmath} % Required for equations

\begin{document}

\title{Domain Knowledge-Enhanced LLMs for Fraud and Concept Drift Detection} 

% Domain Knowledge-Enhanced Dual-LLM Framework for Fraud Detection and Concept Drift Classification in Online Conversations
%\begin{comment}

\author{Ali Şenol, Garima Agrawal, and Huan Liu

\thanks{Ali Şenol is with the School of Computing and Augmented Intelligence (SCAI), Arizona State University (ASU), Tempe, AZ 85281, USA, and the Department of Computer Engineering, Tarsus University, Takbaş M. Kartaltepe S., Tarsus 33400, Mersin, Türkiye (e-mail: asenol@asu.edu; alisenol@tarsus.edu.tr).}
\thanks{Garima Agrawal is with School of Computing and Augmented Intelligence (SCAI), Arizona State University (ASU), Minerva CQ and HumaConn AI Consulting, AZ, USA (e-mail: garima@humaconn.com).}
\thanks{Huan Liu is with the School of Computing and Augmented Intelligence (SCAI), Arizona State University, Tempe, AZ 85281, USA (e-mail: huanliu@asu.edu).}
}
%\end{comment}

\maketitle
\thispagestyle{empty}

\begin{abstract}
Detecting deceptive conversations on dynamic platforms is increasingly difficult due to evolving language patterns and \textit{Concept Drift (CD)}—i.e., semantic or topical shifts that alter the context or intent of interactions over time. These shifts can obscure malicious intent or mimic normal dialogue, making accurate classification challenging. While Large Language Models (LLMs) show strong performance in natural language tasks, they often struggle with contextual ambiguity and hallucinations in risk-sensitive scenarios. To address these challenges, we present a \textbf{Domain Knowledge (DK)-Enhanced LLM framework} that integrates pretrained LLMs with structured, task-specific insights to perform \textit{fraud and concept drift detection}. The proposed architecture consists of three main components: (1) a DK-LLM module to detect fake or deceptive conversations; (2) a drift detection unit (OCDD) to determine whether a semantic shift has occurred; and (3) a second DK-LLM module to classify the drift as either \textit{benign} or \textit{fraudulent}. We first validate the value of domain knowledge using a fake review dataset and then apply our full framework to \textbf{SEConvo}, a multiturn dialogue dataset that includes various types of fraud and spam attacks. Results show that our system detects fake conversations with high accuracy and effectively classifies the nature of drift. Guided by structured prompts, the LLaMA-based implementation achieves \textbf{98\% classification accuracy}. Comparative studies against zero-shot baselines demonstrate that incorporating domain knowledge and drift awareness significantly improves performance, interpretability, and robustness in high-stakes NLP applications.
\end{abstract}

% Detecting deceptive and evolving conversations in real time remains a major challenge, especially on dynamic online platforms prone to concept drift, where conversation topics shift rapidly. Although large language models (LLMs) excel at natural language understanding, they often struggle in risk-sensitive applications due to contextual ambiguity and hallucinations. In this work, we introduce a Domain Knowledge-Enhanced LLM framework that augments pretrained LLMs with structured, context-aware signals derived from labeled datasets. We first demonstrate the effectiveness of our solution in identifying fake reviews and then extend the approach to fraud detection on SEConvo - a benchmark of multi-turn social engineering chats. When guided by structured insights, the LLaMA model achieves a classification accuracy of 98\%, outperforming zero-shot LLMs and traditional classifiers. Our findings show that incorporating domain-grounded cues significantly improves both accuracy and interpretability, enhancing the reliability of LLMs for real-time use in risk-sensitive scenarios.


\section{Introduction}

Detecting fraudulent conversations on digital platforms is essential for ensuring user safety and preserving trust. However, this task is complicated by the evolving nature of language and the presence of natural topic shifts, commonly referred to as \textit{concept drift}, which can obscure or mimic deceptive behavior. While prior approaches have attempted to detect fraud in online dialogues, they often overlook the dynamic interplay between deception and drift. 

We argue that fraud detection and concept drift should not be treated as isolated tasks. In real-world interactions, fraudulent conversations frequently involve subtle shifts in tone, topic, or intent that resemble benign transitions. Failing to account for these shifts may result in false positives, while detecting drift without contextual reasoning may overlook emerging deception. This paper proposes a unified framework to jointly model these phenomena, aiming not to disentangle them, but to enhance detection accuracy by leveraging their interdependence.

Traditional machine learning (ML) techniques have been widely applied to text classification tasks~\cite{info2019textsurvey,mdpi2023textclass,jcco2023bert}. While effective in static environments, these methods struggle in dynamic environments where linguistic patterns and adversarial tactics continuously evolve. Most ML models rely on fixed features and supervised training, making them ill-equipped to adapt to the fluid, ambiguous nature of multi-turn conversations~\cite{palanivinayagam2023twenty,zhang2021adaptive}. In particular, they often fail to distinguish between genuine topical shifts and manipulative intent—both manifestations of concept drift. To ground the discussion that follows, we clarify the core terminology in Table~\ref{terminology}.

LLMs have emerged as powerful tools for language understanding, generation, and contextual reasoning. Their ability to model long-range dependencies makes them promising for tasks such as fraud detection~\cite{singh2025advanced, jiang2024detecting}. However, their use in high-stakes applications remains limited due to issues like hallucinations, contextual ambiguity, and lack of domain adaptation~\cite{sciencedirect2024llmsecurity, sciencedirect2024featuredrift}. These limitations reduce their reliability in detecting nuanced patterns of deception or drift in conversations.

To address these challenges, we investigate whether incorporating domain-specific knowledge into LLM-based pipelines can enhance their accuracy and robustness. Building on prior efforts in ensemble classifiers and modular LLM architectures, we propose a domain-knowledge-enhanced, end-to-end framework that performs three interlinked tasks: detecting deceptive conversations, identifying concept drift, and classifying drift as benign or adversarial in streaming dialogue data.

The key contributions of this work are summarized as follows:

\begin{itemize}
    \item \textbf{Domain Knowledge-Enhanced Dual-LLM Framework:} We present a modular framework that integrates domain-specific knowledge into a dual-LLM architecture to detect fraudulent conversations, identify concept drift, and classify the drift as either benign or adversarial—tasks typically addressed in isolation.
    
    \item \textbf{Three-Stage Detection Pipeline:} Our system comprises (1) a DK-LLM for detecting fake or deceptive conversations, (2) an OCDD-based drift detector for identifying semantic shifts, and (3) a second DK-LLM to classify the nature of the drift. This structured pipeline enhances robustness, modularity, and interpretability in dynamic, multi-turn streaming dialogues.
    
    \item \textbf{Impact of Structured Domain Prompts:} We demonstrate that embedding domain-specific knowledge into prompt design significantly boosts the performance and explainability of LLMs across both synthetic (fake reviews) and real-world (SEConvo) datasets. Our results underscore the importance of domain-aware prompting for high-stakes classification.
    
    \item \textbf{Comprehensive Benchmarking Across LLMs:} We conduct an extensive evaluation of multiple LLM backbones (ChatGPT, Claude, LLaMA, DeepSeek) under both zero-shot and domain-guided settings, highlighting the differential benefits of domain knowledge and identifying optimal configurations for fraud and drift detection.
\end{itemize}

The rest of the paper is organized as follows. Section~\ref{relatedwork} reviews related work on fraud detection, concept drift modeling, and the use of LLMs in dynamic settings. Section ~\ref{methodology} presents the motivation and objectives of our study, while Section~\ref{fake} evaluates the effectiveness of domain knowledge in enhancing LLM performance for detecting fake reviews. Section~\ref{framework} introduces the proposed DK-enhanced Dual-LLM framework, detailing the system architecture and concept drift detection methodology. Section~\ref{discussion} presents the experimental results and comparative analyses across multiple LLMs. Finally, Section~\ref{conclusion} concludes the paper and outlines directions for future research.

\section{Related Work} \label{relatedwork}

The detection of fraudulent conversations and the handling of concept drift in online platforms have garnered significant attention in recent years. Early approaches to fraud detection relied heavily on traditional machine learning (ML) techniques, such as classifiers k-Nearest Neighbors (kNN) \cite{guo2003knn}, Support Vector Machines (SVM) \cite{hearst1998support}, Naive Bayes \cite{john1995estimating}, Random Forest \cite{breiman2001random}, and XGBoost \cite{chen2016xgboost}, which were mainly applied to text classification tasks~\cite{info2019textsurvey, li2022review, mienye2024survey}. Kowsari et al.~\cite{info2019textsurvey} provide a comprehensive survey of these algorithms, highlighting their effectiveness in static environments.

However, a critical limitation of these methods is their reliance on pre-defined features and labeled datasets, making them less adaptable to the fluid and ambiguous nature of online conversations~\cite{palanivinayagam2023twenty, zhang2021adaptive}. As noted by Palanivinayagam et al.~\cite{palanivinayagam2023twenty}, these models often struggle to differentiate between genuine topic changes and actual fraud, both of which are manifestations of concept drift.

To address the challenges posed by evolving data streams, researchers have explored adaptive online incremental learning techniques. Zhang et al.~\cite{zhang2021adaptive} proposed an online incremental adaptive learning method to evolve data streams, with the aim of improving the adaptability of the model. However, these methods often require continuous retraining, which can be computationally expensive and may not be scalable in evolving data stream settings. In addition, they do not explicitly address the issue of interpretability, making it difficult to understand why a particular conversation was flagged as fraudulent.

Concept drift which is defined as a shift in the statistical properties of a target variable over time, poses a critical challenge for online classification systems. Multiple studies have proposed drift adaptation methods to address evolving data distributions~\cite{gama2014survey, zliobaite2010overview, lu2019learning}. These include windowing, ensemble learning, and incremental adaptation techniques that dynamically retrain models in response to detected shifts.

One-Class-based approaches, such as OCDD~\cite{gozuaccik2021concept}, use statistical modeling of normal behavior to detect deviations without requiring labeled drift data. While these techniques enhance adaptability, many still suffer from limited interpretability and high retraining costs in dynamic environments. More recent surveys~\cite{raza2022driftreview} provide systematic taxonomies of drift types and detection algorithms, emphasizing the trade-offs between responsiveness and computational efficiency.

In recent years, LLMs have emerged as powerful tools for natural language understanding and generation. Their capacity to model context at scale makes them appealing for tasks like fraud detection~\cite{singh2025advanced, openai2023gpt}. Singh et al.~\cite{singh2025advanced} explored the use of RAG-based LLMs for advanced real-time fraud detection. However, LLMs also face critical limitations, including susceptibility to hallucinations, contextual ambiguity, and lack of domain adaptation. Yao et al.~\cite{sciencedirect2024llmsecurity} provide a survey on the security and privacy challenges associated with LLMs, highlighting their vulnerability to adversarial attacks and privacy breaches. Furthermore, the `black box' nature of many LLMs makes it difficult to interpret their decision-making processes, which is a significant concern in high-stakes applications.

Building upon these efforts, our work investigates whether incorporating domain-specific knowledge into LLM-based pipelines can enhance their performance in detecting fraudulent conversations and classifying different types of concept drift. We propose a domain knowledge-enhanced LLM framework that addresses the limitations of existing approaches by combining the strengths of LLMs with structured domain insights.

\section{Methodological Overview}\label{methodology}

\subsection{Motivation and Objectives}

The growing prevalence of deceptive behavior on online platforms—including isolated fake reviews and multi-turn fraudulent conversations—poses a serious threat to digital trust and user safety. While existing systems often treat fake content detection and concept drift modeling as separate challenges, they frequently overlook the linguistic and behavioral overlap between these forms of deception. This fragmentation limits generalizability and reduces interpretability, especially in dynamic contexts.

Further complicating the problem is the presence of \textit{concept drift}, where natural topic shifts may be incorrectly flagged as fraudulent, while adversarial pivots may go unnoticed. These challenges motivate the development of a unified, domain-aware approach to detecting deception and drift in both static and streaming dialogue data.

This study is guided by three core objectives: (1) to evaluate whether domain-specific knowledge improves the performance of LLMs in high-stakes natural language tasks, (2) to examine the role of LLMs in detecting both static (e.g., fake reviews) and dynamic (e.g., evolving fraud) deception, and (3) to design a joint framework that handles fraud detection and concept drift classification.

To achieve these goals, we adopt a two-phase methodology:

\begin{itemize}
    \item \textbf{Phase I – Fake Review Detection in Static Settings:} We assess whether domain-specific prompting enhances the accuracy of LLMs in classifying fake reviews. This controlled setting enables us to evaluate the model's ability to internalize structured knowledge and recognize deceptive linguistic patterns. Insights from this phase inform the prompt design and LLM behavior used in downstream tasks.

    \item \textbf{Phase II – Fraud and Drift Detection in Streaming Dialogues:} Building on Phase I, we extend the framework to multi-turn conversations. The proposed pipeline: (a) uses a DK-enhanced LLM to flag deceptive interactions, (b) applies a One-Class Concept Drift Detector (OCDD) to identify semantic shifts, and (c) leverages a second domain-aware LLM to classify each drift as benign (e.g., off-topic, spam) or adversarial (e.g., phishing, manipulation).
\end{itemize}

This staged methodology—from static pattern recognition to dynamic conversation modeling—demonstrates the feasibility and advantages of domain-informed LLMs. Figure~\ref{ConversationDetectionwithLLMModels} provides an overview of this transition, illustrating the interplay between static review analysis and multi-turn fraud detection using our modular pipeline. The framework supports accurate, interpretable fraud detection suitable for deployment in high-stakes domains such as cybersecurity, online marketplaces, and customer service platforms.

\subsection{Preliminaries}

To ground our methodology, we define several key concepts central to the problem space. These distinctions guide both the structure and evaluation of our framework and are summarized in Table~\ref{terminology}.

\textbf{Fake reviews} are fabricated user-generated texts—typically found on platforms like Yelp, Amazon, or TripAdvisor—that misrepresent a product, service, or experience. These are generally short, single-turn entries intended to manipulate public perception. In our framework, fake reviews are detected based on linguistic, syntactic, and semantic irregularities.

\textbf{Fraudulent conversations}, by contrast, involve multi-turn interactions where an adversary seeks to manipulate the user. These dialogues often evolve over time and aim to manipulate the target through tactics such as phishing, impersonation, or baiting. Detecting such interactions requires temporal modeling and pragmatic reasoning.

\textbf{Deceptive behavior} serves as an umbrella term encompassing both fake reviews and fraudulent conversations. It involves the intentional dissemination of misleading or false information, whether in static or interactive form. Our framework models deception as a spectrum of adversarial intent, combining content- and behavior-based cues.

\textbf{Concept drift} refers to a shift in the semantic or statistical properties of a dialogue over time. Such drift may be benign (e.g., topic change) or adversarial (e.g., manipulative pivot). In fraud detection, failing to distinguish between the two can result in false positives or undetected threats. Our system addresses this by combining statistical drift detection (via OCDD) with LLM-based semantic classification.

These concepts directly inform the architecture and logic of our detection pipeline. Specifically:
\begin{itemize}
    \item \textbf{Fake reviews} are first identified using a domain-informed LLM classifier trained on static linguistic patterns.
    \item \textbf{Multi-turn conversations} are evaluated for anomalous behavior indicative of attempts to manipulate the user.
    \item \textbf{Semantic shifts} are flagged using OCDD to identify potential concept drift.
    \item \textbf{LLM-based reasoning} is used to interpret drift and classify it as benign or adversarial.
\end{itemize}

By operationalizing these distinctions, our framework captures the nuanced interplay between static deception, interactive manipulation, and evolving conversation context, thereby improving both the accuracy and interpretability of detection systems.

\begin{table*}[ht]
\centering
\caption{Taxonomy of Concepts Applied in the Study}
\label{terminology}
\begin{tabular}{@{}p{3.5cm}p{12cm}@{}}
\toprule
\textbf{Term} & \textbf{Definition} \\
\midrule
\textbf{Fake Review} & Fabricated user-generated content that misrepresents a product, service, or experience—typically single-turn and found on platforms like Yelp, Amazon, or TripAdvisor. Used to unfairly boost or damage reputations. \\
\textbf{Fraudulent Conversation} & Multi-turn dialogue in which an adversary seeks to manipulate the interlocutor, often using various techniques such as phishing, impersonation, baiting, or pretexting. \\
\textbf{Deceptive Behavior} & A unifying category that includes both fake reviews and fraudulent conversations. Characterized by the intentional presentation of misleading or false information, whether in static or interactive form. \\
\textbf{Concept Drift} & A shift in the statistical or semantic properties of a conversation over time. May be benign (e.g., topic change) or adversarial (e.g., manipulation), complicating fraud detection. \\
\bottomrule
\end{tabular}
\end{table*}

\section{Phase I: Fake Review Detection with Domain Knowledge-Enhanced LLMs}
\label{fake}

Detecting fake reviews—fabricated user-generated content designed to mislead readers—remains a persistent challenge for platforms that rely on crowd-sourced feedback, such as Yelp, Amazon, and TripAdvisor. These reviews often feature exaggerated praise, vague sentiment, or unnatural linguistic structures, making them difficult to detect using conventional machine learning techniques. While traditional classifiers trained on labeled data offer some utility, they frequently lack generalizability and adaptability, particularly in the presence of contextual and stylistic variability.

LLMs provide strong semantic reasoning and contextual understanding, making them well-suited for nuanced fake review detection. However, in zero-shot or general-purpose settings, their performance is often limited by a lack of domain-specific awareness. This can lead to misclassifications or hallucinations when input deviates from the pretraining distribution.

To address this gap, we explore whether infusing domain-specific knowledge—derived from labeled examples and curated linguistic patterns—into the LLM inference process can enhance both accuracy and interpretability. By structuring prompts to reflect known characteristics of deceptive reviews, we guide the model's attention to contextually relevant features while discouraging reliance on superficial or irrelevant cues.

This phase serves both as a proof of concept for domain-informed prompting and as a foundational stage for the more complex fraud detection system developed in Phase II. Our experiments focus on the Yelp dataset, a popular platform where user-generated reviews are susceptible to manipulation by bots or malicious actors. Accurate detection of such content is essential to preserving platform integrity and user trust.

\subsection{Domain Knowledge Pattern Discovery}

To identify common patterns and linguistic markers of fake reviews, we supplied the LLM with a curated set of pre-labeled examples from the Yelp dataset, as illustrated in Figure~\ref{FakeReviewDetectionwithDK}.

\begin{strip}
\begin{tcolorbox}[
    colback=gray!5,
    colframe=black!80,
    boxrule=0.5mm,
    arc=2mm,
    enhanced,
    width=\textwidth,
    title=\textbf{Pattern Extraction in Fake Reviews (Yelp) Dataset},
    fonttitle=\bfseries,
    center title
]
\centering
\includegraphics[width=0.8\textwidth]{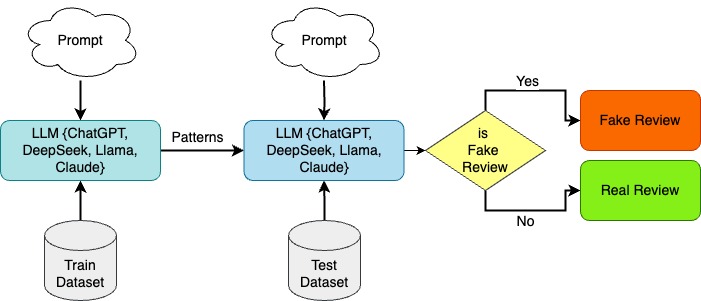}
\captionof{figure}{The workflow employed to discover linguistic patterns and categorical features in the Yelp dataset using a user-defined prompt, designed to guide the LLM in identifying characteristics commonly associated with fake reviews.}
\label{FakeReviewDetectionwithDK}
\end{tcolorbox}
\end{strip}

The domain knowledge used to guide LLM prompting was derived through a hybrid process combining manual annotation of labeled data with exploratory outputs from an initial prompt-based analysis. Specific examples of exaggerated claims, vague statements, and inconsistent sentiments were extracted and categorized to create a pattern library used in downstream prompt templates.

By analyzing these labeled examples, the model was able to identify and articulate distinctive linguistic and stylistic features frequently found in fake reviews. These include exaggerated language, overly generic expressions, inconsistent sentiment, and unnatural phrasing, as shown in Figure~\ref{spam_vs_fraud_result}. This domain-specific insight serves as the foundation for improving the model’s ability to distinguish deceptive content from authentic reviews in downstream classification tasks.

\begin{figure*}[ht]
\centering
\begin{tcolorbox}[colframe=black!60, colback=red!3!white, title= Domain Knowledge Discovered from Yelp Dataset, width=0.95\textwidth]

\textbf{\textit{Identified Fake Review Categories:}}

\begin{itemize}
    \item \textbf{Overly Positive Reviews:} Exaggerated praise with superlatives.  
    \textit{e.g., "Best experience ever! Everything was perfect!"}
    
    \item \textbf{Copy-Pasted Content:} Reused text across platforms.  
    \textit{e.g., Generic review found verbatim in multiple places.}
    
    \item \textbf{Fake Expert Claims:} Posing as professionals without credibility.  
    \textit{e.g., "As a food critic, I guarantee..."}
    
    \item \textbf{Buzzword Overuse:} Trendy terms lacking substance.  
    \textit{e.g., "Instagrammable, artisanal, obsessed!"}
    
    \item \textbf{Unnatural Language:} Awkward or overly formal phrasing.  
    \textit{e.g., "I am pleased to report..."}
    
    \item \textbf{Inconsistent Sentiment:} Mixed praise and criticism.  
    \textit{e.g., "Great food but terrible experience."}
    
    \item \textbf{Suspicious Timing:} Multiple reviews in short bursts.  
    \textit{e.g., Clustered 5-star reviews within 24 hours.}
    
    \item \textbf{Lack of Details:} Vague statements with no specifics.  
    \textit{e.g., "It was good. I recommend it."}
\end{itemize}

\end{tcolorbox}
\caption{An example of discovered fake review patterns and categories in the Yelp dataset.}
\label{spam_vs_fraud_result}
\end{figure*}

\subsection{Dataset}

We utilize the \textit{Yelp Fake Review Dataset}~\cite{yelp2019dataset} to support pattern discovery and classification of deceptive content. The dataset comprises 20,000 reviews, including 15,000 genuine and 5,000 fabricated entries. A summary of the class distribution is shown in Table~\ref{yelp}.

\begin{table}[h]
\centering
\caption{Summary of Yelp Dataset}\label{yelp}
\begin{tabular}{@{}lcc@{}}
\toprule
\textbf{Label} & \textbf{Count} & \textbf{Percentage} \\
\midrule
Real  & 15,000 & 75\%  \\
Fake  & 5,000  & 25\%  \\
\midrule
Total & 20,000 &       \\
\bottomrule
\end{tabular}
\end{table}

\subsection{Evaluation Design}

To assess the impact of domain knowledge on LLM performance, we implemented a two-phase evaluation framework. In the first phase, each model was prompted to classify reviews as real or fake without any domain-specific guidance, establishing a zero-shot baseline. In the second phase, we introduced domain-specific knowledge—comprising the linguistic and behavioral features identified earlier—into the models via structured prompts or fine-tuning and re-evaluated their performance on the same classification task (see Figure~\ref{FakeReviewDetectionwithDK}).

We conducted this evaluation across several state-of-the-art LLMs, including ChatGPT, DeepSeek, LLaMA, and Claude. Each model was tested under two conditions: (1) without domain knowledge (\textbf{\textit{DK}}) and (2) with domain knowledge (\textbf{\textit{DK}}) integrated. This setup allowed for a direct comparison of how domain-specific insights influence classification accuracy and model adaptability.

\subsection{Results and Comparative Analysis}

Table~\ref{comparisonfakes} presents the classification accuracy of the evaluated LLMs with and without domain knowledge integration. The results underscore the value of incorporating domain-specific cues, while also revealing variation in how different models respond to structured prompting.

Claude demonstrated the highest performance, improving from 87\% to 95\% accuracy with domain knowledge. DeepSeek also showed a substantial gain, increasing from 56\% to 90\%. In contrast, ChatGPT displayed marginal improvement (54\% to 56\%), suggesting either prompt design limitations or inherent insensitivity to domain cues in this task setting. LLaMA exhibited a modest gain, rising from 52\% to 58\%.

These findings highlight the importance of model selection, prompt engineering, and domain adaptation in enhancing the reliability of LLMs for deception detection tasks.

\begin{table*}[h]
\centering
\caption{Comparison of prediction accuracy of LLM models on the Yelp fake review dataset}
\label{comparisonfakes}
\begin{tabular}{@{}lcc@{}}
\toprule
\textbf{LLM Model} & \textbf{With DK (\%)} & \textbf{Without DK (\%)} \\
\midrule
LLaMA     & 58  & 52  \\
DeepSeek  & 90  & 56  \\
ChatGPT   & 56  & 54  \\
\textbf{Claude}   & \textbf{95}  & \textbf{87}  \\
\bottomrule
\end{tabular}
\end{table*}

\section{Phase II: Dual-LLM Framework for Fraud and Concept Drift Detection}
\label{framework}

\subsection{Motivation and Objectives}

While traditional classifiers and even standalone LLMs have shown promise in detecting spam and fraudulent conversations, they face significant limitations in dynamic environments—particularly when deception is embedded in subtle shifts in tone or topic. Existing methods often conflate benign concept drift with malicious behavior, leading to false positives or missed detections.

This motivates the development of an adaptive and interpretable architecture that can jointly detect deception and analyze semantic drift in streaming conversations. Specifically, we propose a domain knowledge-enhanced Dual-LLM framework with the following objectives:

\begin{itemize}
    \item \textbf{Fraud Detection:} Leverage a DK-enhanced LLM to classify whether a multi-turn conversation contains potentially deceptive or malicious content.

    \item \textbf{Concept Drift Detection:} Integrate a lightweight statistical detector (OCDD) to identify shifts in topic or tone that deviate from normal dialogue patterns.

    \item \textbf{Drift Type Classification:} Use a second DK-enhanced LLM to interpret the detected drift and classify it as either benign (e.g., spam) or adversarial (e.g., phishing or manipulation).

    \item \textbf{Comparative Analysis:} Evaluate the framework—both with and without DK—against traditional ensemble-based models in terms of accuracy, interpretability, and readiness for practical deployment.
\end{itemize}

Together, these objectives contribute to a comprehensive strategy for detecting and interpreting deceptive behavior in real-world conversational data streams.

\subsection{Integration of Domain Knowledge}

A core component of the proposed framework is the integration of domain-specific knowledge into LLM inference. This is achieved by embedding well-defined linguistic cues and behavioral signals—frequently observed in manipulative conversations—into structured prompts. Key domain-specific cues include:

\begin{itemize}
    \item Use of excessive flattery or persuasive language
    \item Requests for personally identifiable information (PII)
    \item Expressions of urgency related to payment or access
    \item Abrupt or unnatural shifts in tone or topic
\end{itemize}

These cues guide the model's attention toward contextually relevant signals, improving its ability to differentiate between benign and adversarial interactions. The domain-guided prompting strategy serves as a soft constraint, improving interpretability without sacrificing generative flexibility—especially critical in high-stakes scenarios.

\subsection{Concept Drift Detection with OCDD}

To detect concept drift in flagged conversations, we utilize the \textbf{One-Class Concept Drift Detector (OCDD)}~\cite{gozuaccik2021concept}. OCDD is trained exclusively on non-malicious (normal) conversations using a one-class Support Vector Machine (SVM) approach. Once trained, it continuously monitors streaming dialogues for deviations from learned patterns.

If a drift is detected, the conversation is passed to a second LLM (LLM-2), which applies domain-informed reasoning to classify the drift as either benign or adversarial. This modular approach ensures that statistical anomaly detection (via OCDD) is complemented by semantic interpretation (via LLM-2), enabling both sensitivity and contextual precision.

\subsection{Architecture Overview}

As shown in Figure~\ref{ConversationDetectionwithLLMModels}, the proposed architecture adopts a dual-phase approach to fraud and drift detection in conversational data. Conversations flagged as potentially deceptive are passed to OCDD to identify any semantic drift. If drift is detected, the conversation is routed to \textbf{LLM-2}, which classifies the drift as benign (e.g., spam) or malicious (e.g., phishing or manipulation). LLM-2 then delivers a final assessment of the conversation’s authenticity.

An example output from LLM-2 is shown in Figure~\ref{conversation_result}, illustrating its reasoning process. This staged framework enables robust handling of complex multi-turn scenarios, where deception may emerge gradually. By separating anomaly detection, drift recognition, and semantic interpretation, the system reduces false positives and enhances modularity.

The architecture also supports flexible integration of multiple LLMs and detection modules. This design allows individual components to be updated or swapped without affecting the overall workflow. The hybrid approach—statistical drift detection combined with LLM-based reasoning—yields both high predictive accuracy and interpretability, aligning with the needs of cybersecurity and fraud detection applications.

\begin{strip}
\begin{tcolorbox}[
    colback=gray!5,
    colframe=black!80,
    boxrule=0.5mm,
    arc=2mm,
    enhanced,
    width=\textwidth,
    title=\textbf{Workflow of Proposed Framework},
    fonttitle=\bfseries,
    center title
]
\centering
\includegraphics[width=1\textwidth]{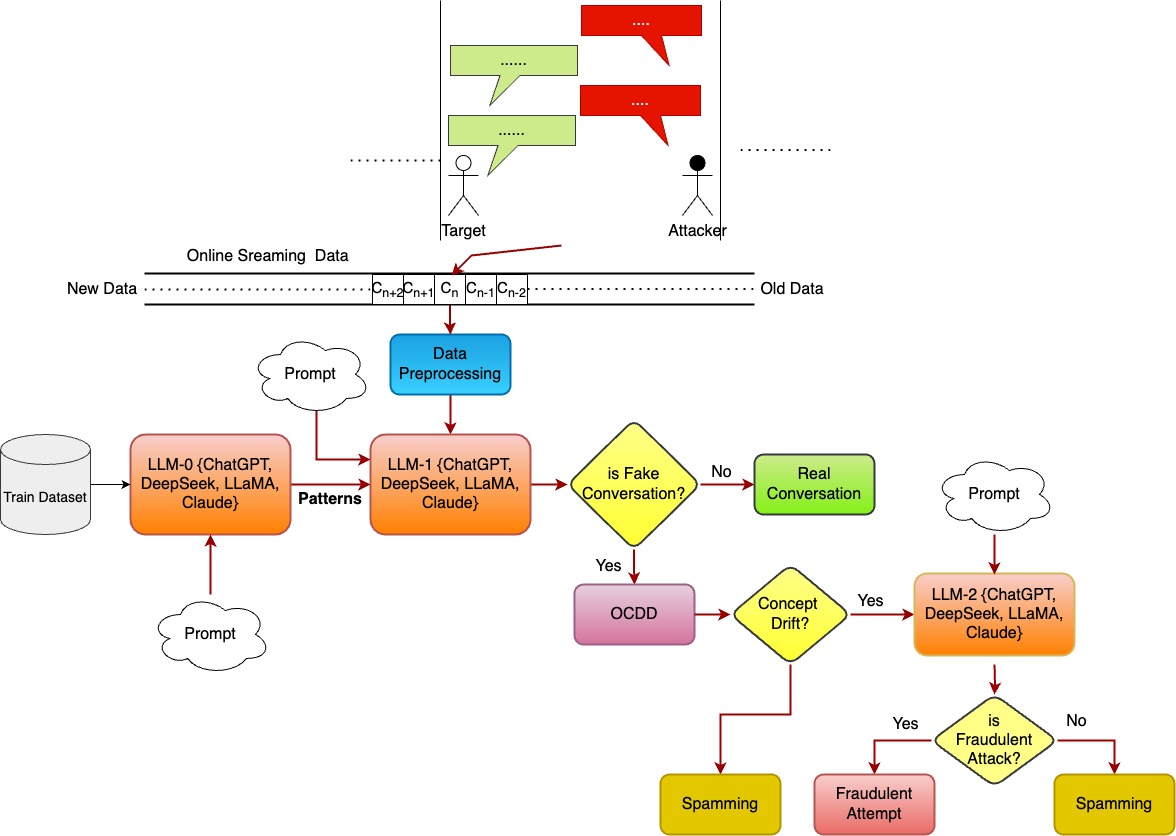}
\captionof{figure}{Workflow of the proposed framework for detecting fraudulent conversations using domain-specific knowledge. The system employs a dual-LLM architecture to identify deceptive interactions, detect semantic shifts, and classify concept drifts as either benign or adversarial.}
\label{ConversationDetectionwithLLMModels}
\end{tcolorbox}
\end{strip}

\begin{figure*}[ht]
\centering
\begin{tcolorbox}[colframe=black!60, colback=red!3!white, title= Sample Response from LLM-2 on Concept Drift Classification , width=0.95\textwidth]

\textit{The conversation appears to be a \textbf{'Fraudulent Attempt'}. The pattern identified is a classic case of phishing whereby \textbf{'Michael'} tries to establish trust, presents an attractive opportunity, and then asks for sensitive personal information and credit card details outside of official channels. It's important to note actions such as asking for the credit card for a nominal fee "to ensure serious collaboration" and activation of the secure portal are common tactics used in fraud. Thus, the type of conversation can be classified as \textbf{'fraudulent'}.}

\end{tcolorbox}
\caption{A sample response from LLM-2 on concept drift classification}
\label{conversation_result}
\end{figure*}

\subsection{Dataset}

Table~\ref{SEConvo} summarizes the SEConvo dataset~\cite{achiam2023gpt} used to train and test the framework. The dataset includes 400 conversations, with 40 used for training and 360 for testing, evenly split across real and fake conversations.

\begin{table}[h]
\centering
\caption{Statistics of the SEConvo dataset used for training and testing}
\begin{tabular}{lcc}
\toprule
\textbf{Label} & \textbf{Train} & \textbf{Test} \\
\midrule
Fake           & 24             & 191           \\
Real           & 16             & 169           \\
Total          & 40             & 360           \\
\bottomrule
\end{tabular}
\label{SEConvo}
\end{table}

\begin{comment}

\begin{table*}[ht]
\centering
\caption{Examples of Benign vs. Adversarial Concept Drift}
\begin{tabular}{|p{3.5cm}|p{4.5cm}|p{4.5cm}|}
\hline
\textbf{Drift Type} & \textbf{Description} & \textbf{Example} \\
\hline
Benign Drift & Natural changes in topic, tone, or expression without deceptive intent & A user switches from a complaint to casual small talk or jokes \\
\hline
Adversarial Drift & Intentional manipulation of topic or language to deceive or extract information & Sudden switch to phishing: “Click this secure link to confirm your password.” \\
\hline
\end{tabular}
\label{drift_examples}
\end{table*}
\end{comment}

\subsection{Results and Comparative Analysis}

The performance of the DK-enhanced Dual-LLM framework was evaluated using the SEConvo dataset, as summarized in Table~\ref{SEConvo}. Table~\ref{comparisonmal} presents a comparison of accuracy, precision, recall, and F1-score for models with and without domain knowledge.

\begin{table*}[h]
\centering
\caption{Comparison of prediction performance of LLM models on Conversation dataset}
\label{comparisonmal}
\begin{tabular}{@{}lcccccccc@{}}
\toprule
\multirow[b]{2}{*}{LLM Model} & \multicolumn{4}{c}{Without DK (\%)} & \multicolumn{4}{c}{With DK (\%)} \\
\cmidrule(lr){2-5} \cmidrule(lr){6-9}
 & Accuracy & Precision & Recall & F1-Score & Accuracy & Precision & Recall & F1-Score \\
\midrule

DeepSeek     &     75    &    80      &     67   &    73   &   76     &    73     &    88    &    80    \\
Claude      &     54    &   58       &   57    &    52   &   67     &     64    &    70    &     66    \\
ChatGPT     &   66      &    80      &  47      &     59   & 71 & 76  & 66 & 71 \\
LLaMA       &     90    &   92       &   88     &    90    &    \textbf{98}    &    \textbf{98}     &    \textbf{98}    &    \textbf{98}   \\
\bottomrule
\end{tabular}
\end{table*}

\textbf{LLaMA} achieved the highest performance across all evaluation metrics, reaching \textbf{98\%} accuracy, precision, recall, and F1-score with domain knowledge. This marks a notable improvement over its 90\% accuracy in the baseline setting. DeepSeek also showed significant gains, particularly in recall (67\% to 88\%), boosting its F1-score to 80\%.

\textbf{Claude} and \textbf{ChatGPT} exhibited more modest improvements. Claude’s F1-score rose from 52\% to 66\%, while ChatGPT increased from 59\% to 71\%. These results underscore the variable sensitivity of LLMs to domain prompting, with LLaMA emerging as the most robust.

For comparison, Şenol et al.~\cite{senol2025joint} used a traditional ensemble-based model combining kNN, SVM, and ImpKmeans~\cite{csenol2024impkmeans} in a majority voting setup. The ensemble model’s performance is shown in Table~\ref{ensemblevsLLM}.

\begin{table*}[h]
\centering
\caption{Comparison of prediction performance of our Dual-LLM framework with DK vs Ensemble model based framework on the SEConvo dataset}
\label{ensemblevsLLM}
\begin{tabular}{@{}lcccc@{}}
\toprule
\textbf{Model} & \textbf{Accuracy (\%)} & \textbf{Precision (\%)} & \textbf{Recall (\%)} & \textbf{F1-Score (\%)} \\
\midrule
Ensemble Model & 82 & 83 & 82 & 82 \\
Ours: DK-Dual-LLM (LLaMA)  &    \textbf{98}    &    \textbf{98}     &    \textbf{98}    &    \textbf{98}   \\
\bottomrule
\end{tabular}
\end{table*}

While the ensemble-based model offered modularity and efficiency, it lacked the semantic depth and contextual reasoning of LLMs. As shown in Table~\ref{ensemblevsLLM}, the DK-enhanced Dual-LLM framework with LLaMA significantly outperforms the ensemble approach across all metrics.

These findings support the viability of DK-enhanced LLMs in fraud detection pipelines where accuracy and interpretability outweigh latency concerns. Future work may explore prompt compression, pruning, or hybrid lightweight architectures to reduce runtime overhead while preserving semantic performance.

\section{Discussion}
\label{discussion}

Our findings demonstrate that incorporating domain-specific knowledge into LLM-based pipelines substantially improves the effectiveness of fraud detection and concept drift classification. Across both static and dynamic deception tasks, domain-guided prompting enhanced not only accuracy but also interpretability and robustness—two qualities critical for deployment in high-stakes, interactive systems.

A key insight from our experiments is the variation in how different LLMs respond to domain-specific prompts. Claude and DeepSeek exhibited strong gains when augmented with domain knowledge, whereas ChatGPT showed more modest improvements. In contrast, LLaMA consistently outperformed all other models across tasks. These disparities likely reflect differences in model architecture, pretraining corpora, and fine-tuning strategies. For instance, models pre-trained on larger conversational datasets may be naturally better at dialog reasoning, while others optimized for general-purpose tasks (e.g., summarization or code generation) may be less responsive to nuanced, structured prompts.

LLaMA’s superior performance may stem from its training emphasis on long-form, dialogue-centric corpora, which appear to enhance its ability to interpret multi-turn interactions and subtle contextual shifts. Additionally, its architecture prioritizes token-level reasoning over high-level abstraction, aligning well with the structured, step-wise prompting strategy adopted in our framework.

The modular design of our architecture further contributes to explainability through clear task decomposition. By separating fraud detection, drift identification, and drift classification, the system supports fine-grained reasoning and reduces the likelihood of compounding hallucination errors. This modularity also promotes semantic precision—particularly important in multi-turn interactions where deceptive behavior often emerges gradually.
Nevertheless, some limitations remain. Despite improvements in contextual reasoning, LLMs still struggle with interpreting sarcasm, cultural idioms, or emotionally ambivalent language. An error analysis revealed that the most frequent failure cases involved exaggerated or ambiguous expressions that were contextually authentic but incorrectly flagged as deceptive. For example, sarcasm and cultural references sometimes led to false positives. Incorporating socio-linguistic features and grounding sentiment interpretation in real-world context may help mitigate these issues.

Moreover, the current system operates in a feed-forward mode, without adaptive correction. Future research could explore human-in-the-loop feedback mechanisms or reinforcement learning strategies to refine decision thresholds dynamically and improve long-term robustness.

Finally, while domain-informed prompting substantially improves detection performance, LLM-based approaches incur higher computational costs compared to traditional models. In high-stakes environments, this trade-off is often acceptable; however, for latency-sensitive applications, optimizing prompts or employing distilled variants of LLMs could offer a more efficient balance between performance and runtime.

% \begin{comment}

% While the inclusion of domain knowledge led to significant performance gains, it also introduced noticeable computational overhead. The observed increase in runtime—particularly for LLaMA and Claude—reflects a fundamental trade-off between reasoning depth and latency. In security-critical applications such as financial fraud detection or national infrastructure monitoring, this trade-off is often justifiable. However, for latency-sensitive use cases like real-time customer service moderation or conversational AI assistants, further model or architecture-level optimizations may be required.
% \end{comment}

% \begin{comment}
% Lastly, although the ensemble-based baseline demonstrated reasonable performance with low computational cost, it lacked the depth and adaptability of LLM-based reasoning. Our results suggest that domain-informed LLMs strike a more favorable balance between accuracy and interpretability—albeit at the cost of increased runtime—making them more suitable for high-precision environments where the cost of false negatives is prohibitive
% \end{comment}

\section{Conclusion}
\label{conclusion}

This paper presented a domain knowledge-enhanced Dual-LLM framework for detecting fake conversations and classifying concept drift in online dialogues. By embedding domain-specific cues into structured prompts and leveraging a modular, two-stage LLM architecture, the proposed approach addresses key limitations of both traditional classifiers and standalone LLMs—particularly in handling contextual ambiguity, mitigating hallucinations, and adapting to evolving language patterns.

Comprehensive evaluations—spanning fake review classification and multi-turn dialogue analysis using the SEConvo dataset—demonstrate the effectiveness of the method. The LLaMA-based implementation achieved a classification accuracy of 98\%, significantly outperforming both zero-shot LLMs and prior ensemble-based models. These results underscore the value of integrating domain knowledge into LLM workflows to achieve high precision and interpretability in complex, real-time detection scenarios.

Future directions include extending the framework to support multilingual and culturally adaptive deception detection, integrating human-in-the-loop or reinforcement learning mechanisms for dynamic refinement, and deploying the system in real-time messaging platforms for proactive fraud prevention. Additionally, enhancing the granularity of drift classification—such as differentiating between promotional, transactional, or personal shifts—could provide deeper insight into the evolution of conversational intent in practical applications.

% \begin{comment}
% While the dual-LLM architecture introduces additional computational overhead, its modularity and semantic robustness make it well-suited for high-stakes environments where accuracy and explainability are paramount. Future research will explore latency reduction strategies to support efficient on-device deployment and expand the system’s adaptability across domains.
% \end{comment}

\section*{Acknowledgments}
This research is supported by TÜBİTAK under the 2219 Postdoctoral Fellowship Program.

\bibliographystyle{IEEEtran}
\bibliography{reference} % Your .bib file name

% Generated by IEEEtran.bst, version: 1.14 (2015/08/26)
\begin{thebibliography}{10}
\providecommand{\url}[1]{#1}
\csname url@samestyle\endcsname
\providecommand{\newblock}{\relax}
\providecommand{\bibinfo}[2]{#2}
\providecommand{\BIBentrySTDinterwordspacing}{\spaceskip=0pt\relax}
\providecommand{\BIBentryALTinterwordstretchfactor}{4}
\providecommand{\BIBentryALTinterwordspacing}{\spaceskip=\fontdimen2\font plus
\BIBentryALTinterwordstretchfactor\fontdimen3\font minus \fontdimen4\font\relax}
\providecommand{\BIBforeignlanguage}[2]{{%
\expandafter\ifx\csname l@#1\endcsname\relax
\typeout{** WARNING: IEEEtran.bst: No hyphenation pattern has been}%
\typeout{** loaded for the language `#1'. Using the pattern for}%
\typeout{** the default language instead.}%
\else
\language=\csname l@#1\endcsname
\fi
#2}}
\providecommand{\BIBdecl}{\relax}
\BIBdecl

\bibitem{info2019textsurvey}
K.~Kowsari, K.~Jafari~Meimandi, M.~Heidarysafa, S.~Mendu, L.~Barnes, and D.~Brown, ``Text classification algorithms: A survey,'' \emph{Information}, vol.~10, no.~4, p. 150, 2019.

\bibitem{mdpi2023textclass}
H.~Allam, L.~Makubvure, B.~Gyamfi, K.~N. Graham, and K.~Akinwolere, ``Text classification: How machine learning is revolutionizing text categorization,'' \emph{Information}, vol. 16(2), no. 130, 2025.

\bibitem{jcco2023bert}
E.~C. Garrido-Merchan, R.~Gozalo-Brizuela, and S.~Gonzalez-Carvajal, ``Comparing bert against traditional machine learning models in text classification,'' \emph{Journal of Computational and Cognitive Engineering}, vol.~2, no.~4, pp. 352--356, 2023.

\bibitem{palanivinayagam2023twenty}
A.~Palanivinayagam, C.~Z. El-Bayeh, and R.~Dama{\v{s}}evi{\v{c}}ius, ``Twenty years of machine-learning-based text classification: A systematic review,'' \emph{Algorithms}, vol.~16, no.~5, p. 236, 2023.

\bibitem{zhang2021adaptive}
S.-s. Zhang, J.-w. Liu, and X.~Zuo, ``Adaptive online incremental learning for evolving data streams,'' \emph{Applied Soft Computing}, vol. 105, p. 107255, 2021.

\bibitem{singh2025advanced}
G.~Singh, P.~Singh, and M.~Singh, ``Advanced real-time fraud detection using rag-based llms,'' \emph{arXiv preprint arXiv:2501.15290}, 2025.

\bibitem{jiang2024detecting}
L.~Jiang, ``Detecting scams using large language models,'' \emph{arXiv preprint arXiv:2402.03147}, 2024.

\bibitem{sciencedirect2024llmsecurity}
Y.~Yao, J.~Duan, K.~Xu, Y.~Cai, Z.~Sun, and Y.~Zhang, ``A survey on large language model (llm) security and privacy: The good, the bad, and the ugly,'' \emph{High-Confidence Computing}, p. 100211, 2024.

\bibitem{sciencedirect2024featuredrift}
F.~Hinder, V.~Vaquet, and B.~Hammer, ``Feature-based analyses of concept drift,'' \emph{Neurocomputing}, vol. 600, p. 127968, 2024.

\bibitem{guo2003knn}
G.~Guo, H.~Wang, D.~Bell, Y.~Bi, and K.~Greer, ``Knn model-based approach in classification,'' in \emph{On The Move to Meaningful Internet Systems 2003: CoopIS, DOA, and ODBASE: OTM Confederated International Conferences, CoopIS, DOA, and ODBASE 2003, Catania, Sicily, Italy, November 3-7, 2003. Proceedings}.\hskip 1em plus 0.5em minus 0.4em\relax Springer, 2003, pp. 986--996.

\bibitem{hearst1998support}
M.~A. Hearst, S.~T. Dumais, E.~Osuna, J.~Platt, and B.~Scholkopf, ``Support vector machines,'' \emph{IEEE Intelligent Systems and their applications}, vol.~13, no.~4, pp. 18--28, 1998.

\bibitem{john1995estimating}
G.~H. John and P.~Langley, ``Estimating continuous distributions in bayesian classifiers,'' in \emph{Proceedings of the Eleventh Conference on Uncertainty in Artificial Intelligence (UAI)}.\hskip 1em plus 0.5em minus 0.4em\relax Morgan Kaufmann, 1995, pp. 338--345.

\bibitem{breiman2001random}
L.~Breiman, ``Random forests,'' \emph{Machine Learning}, vol.~45, no.~1, pp. 5--32, 2001.

\bibitem{chen2016xgboost}
T.~Chen and C.~Guestrin, ``Xgboost: A scalable tree boosting system,'' in \emph{Proceedings of the 22nd ACM SIGKDD International Conference on Knowledge Discovery and Data Mining (KDD)}.\hskip 1em plus 0.5em minus 0.4em\relax ACM, 2016, pp. 785--794.

\bibitem{li2022review}
R.~Li, M.~Liu, D.~Xu, J.~Gao, F.~Wu, and L.~Zhu, ``A review of machine learning algorithms for text classification,'' \emph{Cyber Security}, vol. 226, 2022.

\bibitem{mienye2024survey}
I.~D. Mienye and N.~Jere, ``A survey of decision trees: Concepts, algorithms, and applications,'' \emph{IEEE access}, 2024.

\bibitem{gama2014survey}
J.~Gama, I.~{\v{Z}}liobait{\.e}, A.~Bifet, M.~Pechenizkiy, and A.~Bouchachia, ``A survey on concept drift adaptation,'' \emph{ACM Computing Surveys (CSUR)}, vol.~46, no.~4, pp. 1--37, 2014.

\bibitem{zliobaite2010overview}
I.~{\v{Z}}liobait{\.e}, ``Learning under concept drift: an overview,'' \emph{arXiv preprint arXiv:1010.4784}, 2010.

\bibitem{lu2019learning}
J.~Lu, A.~Liu, F.~Dong, F.~Gu, J.~Gama, and G.~Zhang, ``Learning under concept drift: A review,'' \emph{IEEE transactions on knowledge and data engineering}, vol.~31, no.~12, pp. 2346--2363, 2018.

\bibitem{gozuaccik2021concept}
Ömer Gözüaçık \& Fazli~Can, ``Concept learning using one-class classifiers for implicit drift detection in evolving data streams,'' \emph{Artificial Intelligence Review}, vol.~54, no.~5, pp. 3725--3747, 2021.

\bibitem{raza2022driftreview}
S.~Agrahari and A.~K. Singh, ``Concept drift detection in data stream mining: A literature review,'' \emph{Journal of King Saud University-Computer and Information Sciences}, vol.~34, no.~10, pp. 9523--9540, 2022.

\bibitem{openai2023gpt}
OpenAI, ``Gpt-4 technical report,'' 2023, available at: \url{https://openai.com/research/gpt-4}.

\bibitem{yelp2019dataset}
{Yelp}, ``Yelp open dataset,'' \url{https://www.yelp.com/dataset}, 2019, accessed: 2025-04-22.

\bibitem{achiam2023gpt}
J.~Achiam, S.~Adler, S.~Agarwal, L.~Ahmad, I.~Akkaya, F.~L. Aleman, D.~Almeida, J.~Altenschmidt, S.~Altman, S.~Anadkat \emph{et~al.}, ``Gpt-4 technical report,'' \emph{arXiv preprint arXiv:2303.08774}, 2023.

\bibitem{senol2025joint}
A.~{\c{S}}enol, G.~Agrawal, and H.~Liu, ``Joint detection of fraud and concept drift in online conversations with llm-assisted judgment,'' \emph{arXiv preprint arXiv:https://arxiv.org/abs/2505.07852}, 2025.

\bibitem{csenol2024impkmeans}
A.~Şenol, ``Impkmeans: An improved version of the k-means algorithm, by determining optimum initial centroids, based on multivariate kernel density estimation and kd-tree,'' \emph{Acta Polytechnica Hungarica}, vol.~21, no.~2, pp. 111--131, 2024.

\end{thebibliography}

\clearpage

\appendices

\section{Discovered Patterns and Categories in Fake Reviews (Yelp Dataset)}
\label{appendix:yelp}

\begin{tcolorbox}[colback=lightgraybox, colframe=darkborder, title=Summary of Fake Review Categories Identified Using DK-Enhanced LLM, fonttitle=\bfseries, width=\textwidth, sharp corners=south]

This appendix summarizes the linguistic categories and patterns extracted from the Yelp Fake Review Dataset using structured prompting with a domain knowledge-enhanced LLM.

\vspace{0.5em}
\textbf{Detected Fake Review Categories:}
\begin{itemize}
    \item \textbf{Overly Positive Reviews} — Exaggerated praise using superlatives.\\
    \textit{Example:} ``This is the best restaurant I’ve ever been to!''

    \item \textbf{Copy-Pasted Reviews} — Reused or minimally altered content across platforms.\\
    \textit{Example:} ``The food was delicious and the service friendly.'' (appears identically on multiple sites)

    \item \textbf{Fake Expert Reviews} — Claims of authority with no verifiable background.\\
    \textit{Example:} ``As a food critic, I guarantee this is the best place in town.''

    \item \textbf{Buzzword Overuse} — Excessive use of trendy or hype language.\\
    \textit{Example:} ``So Instagrammable and artisanal. I'm obsessed!''

    \item \textbf{Unnatural Language Patterns} — Robotic or overly formal phrasing.\\
    \textit{Example:} ``I am pleased to report my dining experience was satisfactory.''

    \item \textbf{Inconsistent Sentiment} — Conflicting tone within the same review.\\
    \textit{Example:} ``Great food, terrible service. I'd still recommend it.''

    \item \textbf{Suspicious Timestamps} — Clusters of 5-star reviews in short time spans.\\
    \textit{Example:} Multiple glowing reviews posted within hours.

    \item \textbf{Lack of Specificity} — Vague praise without concrete experiential context.\\
    \textit{Example:} ``It was great. Highly recommended.''
\end{itemize}

\vspace{0.5em}
\textbf{Detection Recommendations:}
\begin{itemize}
    \item \textbf{Linguistic Cues:}
    \begin{itemize}
        \item Flag exaggerated or robotic phrasing.
        \item Detect inconsistent sentiment within a single review.
    \end{itemize}

    \item \textbf{Temporal Patterns:}
    \begin{itemize}
        \item Analyze timestamps for clustered activity.
        \item Flag review bursts during unusual hours.
    \end{itemize}

    \item \textbf{Sentiment Analysis:}
    \begin{itemize}
        \item Identify tonal contradictions or extreme polarity.
        \item Flag reviews exhibiting abrupt emotional shifts.
    \end{itemize}
    
    \item \textbf{Specificity and Contextuality:}
    \begin{itemize}
        \item Detect vague or generic reviews lacking concrete details.
        \item Compare sentiment with peer reviews for deviation.
    \end{itemize}
    
    \item \textbf{Common Red Flags:}
    \begin{itemize}
        \item Overuse of superlatives (e.g., ``amazing,’’ ``best ever’’)
        \item Generic enthusiasm with no substance
        \item Contradictory claims without logical support
    \end{itemize}
    
    \item \textbf{Integration with ML Pipelines:}
        \begin{itemize}
            \item Use labeled review patterns to train supervised models.
            \item Incorporate prompt-derived cues as structured input to LLMs.
        \end{itemize}
\end{itemize}

\vspace{0.5em}
\textit{Incorporating these patterns into both prompt design and classifier training can improve robustness, accuracy, and interpretability of fake review detection pipelines.}

\end{tcolorbox}

\end{document}